\pdfoutput=1 

\documentclass[a4paper,conference]{IEEEtran}
\ifCLASSINFOpdf
\else
\fi
\usepackage{url}

\usepackage{times}
\usepackage{epsfig}
\usepackage{graphicx}
\usepackage{amsmath}
\usepackage{amssymb}

\usepackage{subfigure}
\usepackage{multirow}
\usepackage{amsfonts}
\usepackage{multicol}
\usepackage{booktabs}
\usepackage{algorithm}
\usepackage{algorithmic}
\usepackage{color}
\usepackage{comment}

\newcommand{\etal}{\textit{et al}.}

\begin{document}
%
\title{Text Recognition in Real Scenarios with a Few Labeled Samples}

\author{Jinghuang Lin\textsuperscript{1}\qquad\qquad
	Zhanzhan Cheng\textsuperscript{2}\qquad\qquad
    Fan Bai\textsuperscript{1}\\
	Yi Niu\textsuperscript{2}\qquad\qquad
	Shiliang Pu\textsuperscript{2}\qquad\qquad\qquad
	Shuigeng Zhou\textsuperscript{1}\thanks{Corresponding author.}\\
	\textsuperscript{1}Shanghai Key Lab of Intelligent Information Processing, and School of \\Computer Science, Fudan University, Shanghai 200433, China\\
	\textsuperscript{2}Hikvision Research Institute, China\\
	{\tt\small \{18210240010,fbai19,sgzhou\}@fudan.edu.cn}\\
	{\tt\small11821104@zju.edu.cn}\\
	{\tt\small \{niuyi,pushiliang.hri\}@hikvision.com}
	
}


%


\maketitle

\begin{abstract}
	Scene text recognition (STR) is still a hot research topic in computer vision field due to its various applications.
	Existing works mainly focus on learning a general model with a huge number of synthetic text images to recognize unconstrained scene texts, and have achieved substantial progress. However, these methods are not quite applicable in many real-world scenarios 
	where 1) \textit{high recognition accuracy is required}, while 2) \textit{labeled samples are lacked}.
	To tackle this challenging problem, this paper proposes a few-shot adversarial sequence domain adaptation (FASDA) approach to build sequence adaptation between the synthetic source domain (with \textit{many} synthetic labeled samples) and a specific target domain (with only \textit{some or a few} real labeled samples). This is done by simultaneously learning each character's feature representation with an attention mechanism and establishing the corresponding character-level latent subspace with adversarial learning.
	Our approach can maximize the character-level confusion between the source domain and the target domain, thus achieves the sequence-level adaptation with even a small number of labeled samples in the target domain.
	Extensive experiments on various datasets show that our method
	significantly outperforms the finetuning scheme, and obtains comparable performance to the state-of-the-art STR methods.
	
\end{abstract}

\section{Introduction}
In computer vision area, scene text recognition (STR) has been a hot research topic due to its various applications such as batch number identification in production lines, vehicle license plate recognition in intelligent transportation systems, and container number identification in industrial ports etc.
With the advance of deep learning, many methods ~\cite{bai2018cvpr,cheng2017focus,lee2016recursive,ShiBY17,shi2018aster} have been developed to solve the STR problem. These methods are mainly for training a general model with synthetic datasets (\emph{e.g.} 8-million synthetic data \cite{jaderberg2014synthetic}) and have achieved promising performance on some public benchmarks.
However, they actually do not work well when applied directly to many real STR applications due to 1) the lack of labeled samples and 2) the requirement for very high recognition accuracy (\emph{e.g.}~90+\%, or even higher).
\begin{figure}[!htbp]
	\begin{center}
		\includegraphics[width=0.33\textwidth]{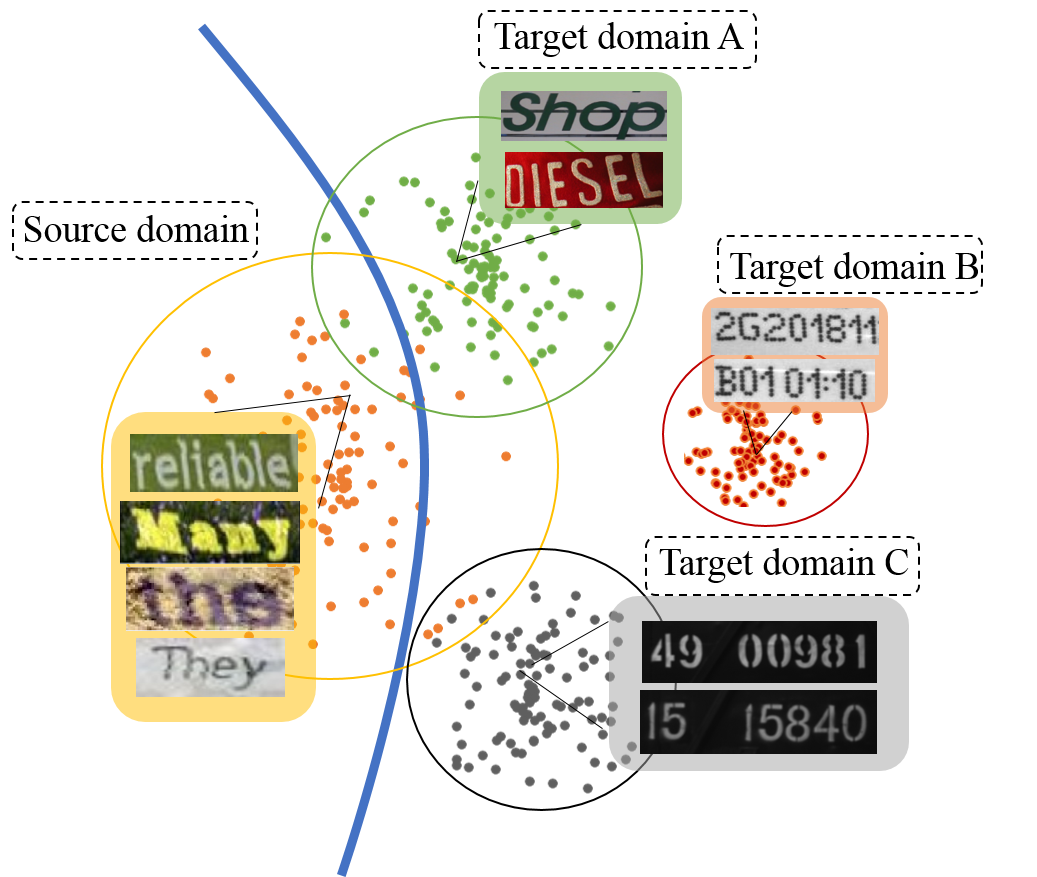}
	\end{center}
	\caption{Illustration of domain adaptation of text images. Here, the \textit{source} domain (left) has a huge number of labeled samples, while each of the three \textit{target} domains \textit{A}, \textit{B} and \textit{C} (right) has relatively much fewer labeled samples.}
	\label{fig:intro}
\end{figure}
There are three possible ways to solve the training sample inadequacy problem:
(1) Reusing some pre-trained model on huge amounts of synthetic data (here also called \textit{source domain}), then \textbf{\textit{finetuning}} the model on some data of the specific application under consideration (here also called \textit{target domain}). Though finetuning is simple to implement, it is prone to overfitting when the labeled samples of the target domain are very limited.
(2) \textbf{\textit{Augmenting}} samples with generative adversarial network techniques such as Cycle-GAN \cite{zhu2017unpaired}. Though the sample generator can augment similar samples and alleviate the sample lack problem to some degree, the styles of generated samples are still uncontrollable. In fact, we have empirically found that such methods are more useful for training a general model as done by most existing works \cite{zhan2018spatial}.
(3) \textbf{\textit{Conducting supervised domain adaptation}} from a \emph{source domain} with a large number of labeled samples to a \emph{target domain} with relatively much fewer labelled samples, as illustrated in Fig.~\ref{fig:intro}.
Actually, some existing supervised few-shot domain adaptation techniques~\cite{csurka2017comprehensive,motiian2017few}, especially the adversarial-based ones, have demonstrated their effectiveness in handling adaptation for classification tasks. However, there is only one work~\cite{zhang2019sequence} that employs domain adaptation for the STR problem~(more details are given in Sec.~\ref{sec:related}).


\begin{figure}[tb]
	\centering
	\subfigure[]{
		\begin{minipage}[t]{0.45\linewidth}
			\centering
			\includegraphics[width=1\textwidth]{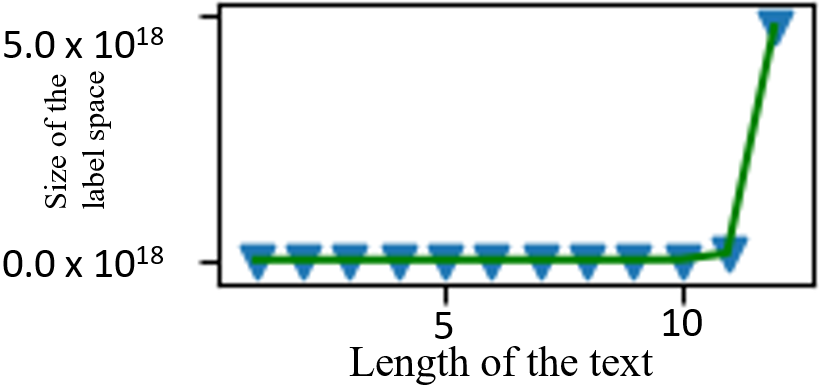}
			\label{fig:CSE}
		\end{minipage}
	}
	\subfigure[]{
		\begin{minipage}[t]{0.419\linewidth}
			\centering
			\includegraphics[width=1\textwidth]{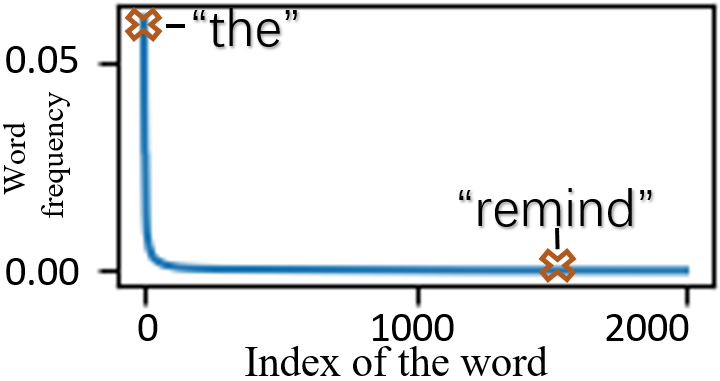}
			\label{fig:LT}
		\end{minipage}
	}
	\centering
	\caption{Challenges facing existing methods for sequence-level domain adaptation: (a) category space explosion, i.e., the possible category number increases exponentially with the sequence length; (b) the long tail phenomenon, i.e., most words (meaningful sequences of characters) appear rarely.
	}
	\label{fig:twotable}
\end{figure}
In this paper, we try to address real STR applications by domain adaptation where we have to face tow challenges: 1) \textit{Category space explosion}. Taking FADA~\cite{motiian2017few} for example, it does image-level adaptation for classification, and each image belongs to one of a limited number of classes. In STR context, an image corresponds to a text sequence, and each sequence is very possibly unique, i.e., with a unique class label (if we must assign it one). So the possible number of classes is proportional to the number of sequences, in other words, exponential to the sequence length, as shown in Fig.~\ref{fig:CSE}.
2) The \textit{long tail} phenomenon. As most sequences are nearly unique, that is, appear rarely. As an example for illustration, Fig.~\ref{fig:LT} shows the distribution of word frequency\footnote{Word frequency is from the statistics of Wikimedia downloaded from \emph{https://dumps.wikimedia.org}.}.
These two challenges make it difficult or even impossible to train a STR model by existing domain adaptation methods.


To circumvent these two problems, an idea is to split each text sequence to a sequence of characters, such that the \emph{sequence domain adaptation} problem is transformed to a \emph{character domain adaptation} problem. However,
explicitly splitting text sequences needs character-level annotations, which will consume huge money and time in practice. A desirable way is implicitly splitting text sequences without relying on character-level geometry information (\emph{e.g.} an image with text ``ABC'' is sequentially annotated as a sequence `A'-`B'-`C' but without position information).

With this idea in mind, here we develop a \textit{few-shot adversarial sequence domain adaptation} (FASDA in short) approach to achieve sequence-level domain confusion. 
First, we represent each character as high-level feature representation with a well-designed attention mechanism (instead of explicitly splitting a text to a character sequence).
Thus, we obtain two groups of character features for pairs of source and target samples. 
Then, we develop a sampling strategy to generate four types of trainable adversarial pairs (instead of typical binary adversarial manner) from the aforementioned two groups of represented characters.
Finally, we apply adversarial learning to the obtained adversarial pairs to achieve domain confusion.
In addition, the attention drift problem \cite{cheng2017focus} is properly addressed to avoid the missing of key character features.
Here, FASDA is proposed as an adversarial sequence learning strategy to guarantee sequence-level domain confusion. 

In summary, our method assembles an attention-based character feature extraction mechanism with the few-shot adversarial sequence learning strategy into a trainable end-to-end framework.
Our contributions are as follows:
1) We propose a few-shot adversarial sequence domain adaptation approach to achieve sequence-level domain confusion by integrating a well-designed attention mechanism with sequence-level adversarial learning strategy into a framework.
2) We implement the framework to fill the performance gap between general STR models and specific STR applications, and show that the framework can be trained end-to-end with much fewer sequence-level annotations.
3) We conduct extensive experiments to show that our method significantly outperforms traditional learning-based schemes such as finetuning, and is competitive with the state-of-the-art STR methods.

\section{Related Work}
\label{sec:related}
\begin{figure*}[tb]
	\begin{center}
		\includegraphics[width=0.89\textwidth]{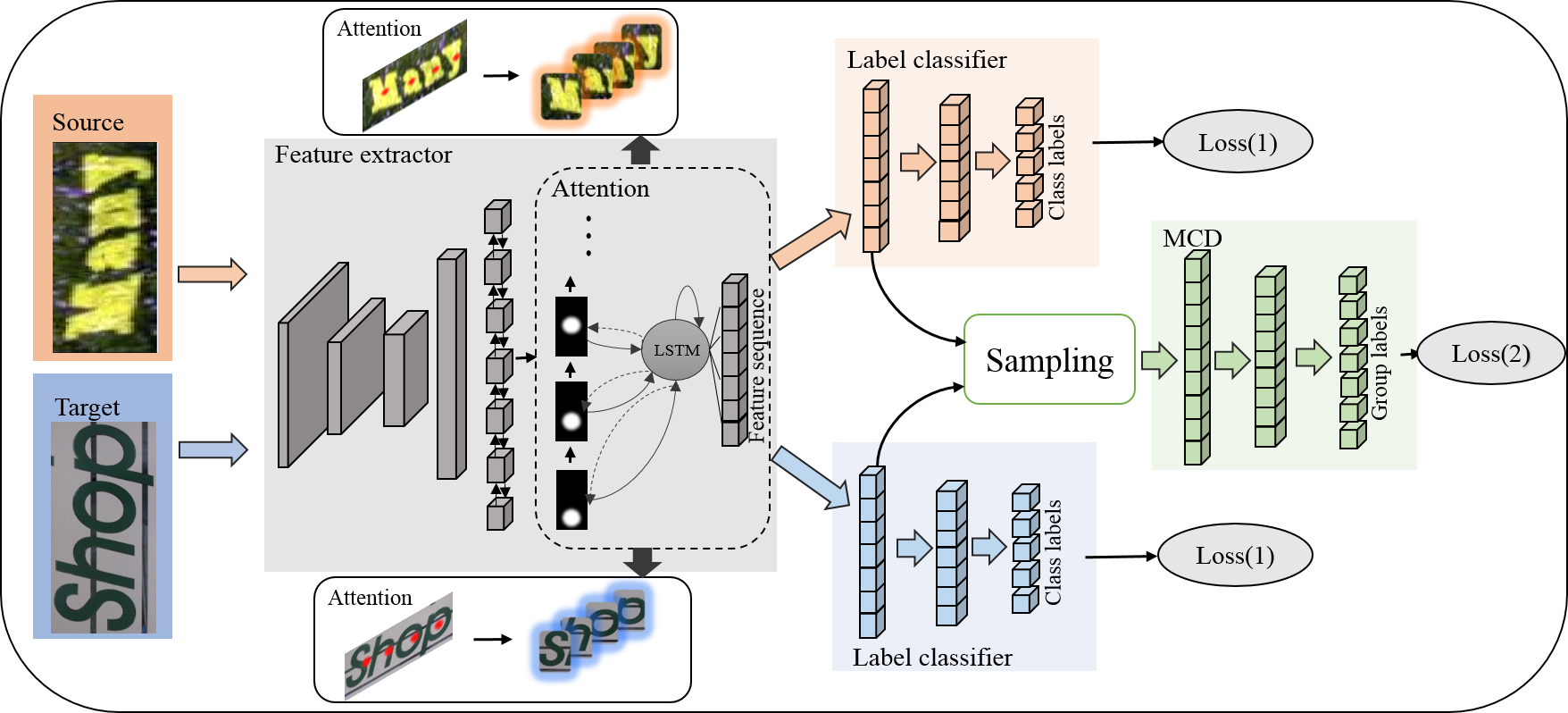}
	\end{center}
	\caption{The framework of FASDA.
		The \emph{Feature extractor} module consists of a CNN-LSTM network and an attention model, which encodes the images into features of characters.
		The \emph{Sampling} module implements the \emph{sampling strategy} that generates feature pairs, and the \emph{MCD} module denotes the MCD function that uses the adversarial learning strategy to do supervised sequence domain adaptation for STR.
		\emph{loss(1)} means the attention loss $\mathcal{L}_{att}$ in Eq.~(\ref{eq:loss_att}), and \emph{loss(2)} indicates the adversarial loss: $\mathcal{L}_{D}$ in Eq.~(\ref{eq:fada-d} ) and $\mathcal{L}_{G}$ in Eq.~(\ref{eq:fada-g}).
	}
	\label{fig:arch}
\end{figure*}
We first briefly review the related works of scene text recognition and domain adaptation, then highlight the differences between our work and the existing methods.

{\bf Scene text recognition}. In recent years, there has been lots of work on scene text recognition\cite{bhunia2019handwriting,li2019show,liao2019mask,long2018textsnake,xie2019aggregation,yang2019symmetry,zhan2019esir,zhang2019sequence,zhan2019ga}. Here we review only the closely related ones and refer the readers to a recent survey 
\cite{lin2019review} for more details.
In early years, deep neural-network based methods were developed for extracting robust visual features.  \cite{jaderberg2016reading,wang2012end} first developed a CNN-based framework for character feature representation, then applied some heuristic rules for characters inference.
Recent works solve this problem as a sequence learning task, where an image is first encoded into a patch sequence and then decoded as a character sequence.
\cite{He2015reading,ShiBY17} proposed end-to-end neural networks that first capture visual feature representation by using CNN or RNN, then the
connectionist temporal classification (CTC) loss is combined with the neural network output to calculate the conditional probability between the predicted and the target sequences. Liu \etal \cite{liu2018synthetically} proposed a synthetically supervised feature learning based method via adversarial learning.
The state-of-the-art of STR is the attention-based methods~\cite{bai2018cvpr,cheng2017focus,cheng2018aon,lee2016recursive,shi2016robust,shi2018aster},
which encode the original images into feature representations with CNN and RNN, and employ a frame-wise loss (\emph{e.g.} cross-entropy loss) to optimize the model.


{\bf Domain adaptation}. 
Domain adaptation aims to generalize a model from the \emph{source} domain to a \emph{target} domain by utilizing available target domain data.
Unsupervised domain adaptation (UDA) does not need any target data labels, but requires a large number of target samples~\cite{ganin2015unsupervised}. However, it is not \textbf{applicable} for our task due to the deadly lack of target samples (\emph{e.g.} the training set size of SVT \cite{wang2011end} is just 257).
Supervised domain adaptation (SDA) performs well when target data are limited but labelled.
Hu \etal \cite{hu2015deep}
applies the marginal Fisher analysis criterion and maximum mean discrepancy to minimize
the distribution difference between the source and target domains.
Gebru \etal \cite{gebru2017fine} designs an attribute and class level classification loss for fine-tuning the target model.
Motiian \etal \cite{motiian2017few} finds a shared subspace for both the source and target
distributions using adversarial learning. 
Recently, several
works~\cite{cao2018partial,dong2018few,Hoffman_cycada2017,kim2017learning,liu2016coupled,tzeng2017adversarial,zou2019consensus} show the ability of adversarial learning in domain adaptation.

{\bf Differences between our work and existing method}.
To the best of our knowledge, there is only one work that explores visual adaptation for STR~\cite{zhang2019sequence}. However, our work differs from this one in at least two aspects: 1) our method utilizes the implicit character annotation information contained in text annotation, it not only aligns the distributions of the source and target domains, but also promotes the semantic alignment of classes. 2) Our method takes advantage of adversarial learning that has been proved effective in domain adaptation.
%
%
And experiments on SVT, IC03, and IC13 show that our method can get better performance. 


\section{Method}

Fig.~\ref{fig:arch} shows the framework of our FASDA method, which consists of two major procedures: 1) weakly-supervised representation of character-level features with attending mechanism while addressing the attention drift issue, 2) few-shot adversarial learning with a specifically designed sampling strategy. 


\subsection{Weakly-supervised Character Feature Representation} \label{sec:attention}
The feature representations of characters can be learnt by the attending mechanism \cite{bahdanau2014neural} that is trained with only sequence-level annotations.

\emph{Attending mechanism}.
Given a text image $\mathcal{I}$ represented as a sequence of feature vectors $({x}_1, ..., {x}_M)$ of length $M$, the attending mechanism is responsible for recursively assigning a scalar value as a weight for $x_j$.
Specifically, the attending weight of the $t$-th character is calculated by
\begin{equation}
\begin{split}
\alpha_{t,j} &= \frac{exp(e_{t,j})}{\sum_{i=1}^{M}exp(e_{t,i})} \\
\end{split}
\label{eq:alpha}
\end{equation}
where $e_{t,j} = w^T tanh(Ws_{t-1} + Vx_j +b)$ is the learnt energy value by simultaneously considering the current feature vector $x_j$ and the hidden state $s_{t-1}$ of the last output, and $W$, $V$ and $b$ are trainable parameters.
With the learnt attending weights $\alpha$, the feature representation $CR_t$ of the $t$-th character can be directly denoted as
\begin{equation}
\begin{split}
CR_t &= \sum_{j=1}^M \alpha_{t,j}x_j.
\end{split}
\label{eq:character-feature1}
\end{equation}
As there exists lexical dependence among characters, it is better to encode such dependence into $CR_t$. This can be done with the long-short term memory (LSTM) network by using the relationship between $CR_t$ and the last character $y_{t-1}$. Thus, we get $CR_t^+$ as follows:
\begin{equation}
\begin{split}
CR_t^+&\overset{def}{=} s_t \\
&=LSTM(y_{t-1},CR_t,s_{t-1}).
\end{split}
\label{eq:character-feature2}
\end{equation}

\begin{figure}[tb]
	\centering
	\subfigure[]{
		\begin{minipage}[t]{0.44\linewidth}
			\centering
			\includegraphics[width=1\textwidth]{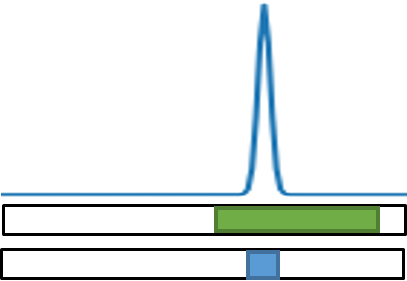}
		\end{minipage}
	}
	\subfigure[]{
		\begin{minipage}[t]{0.44\linewidth}
			\centering
			\includegraphics[width=1\textwidth]{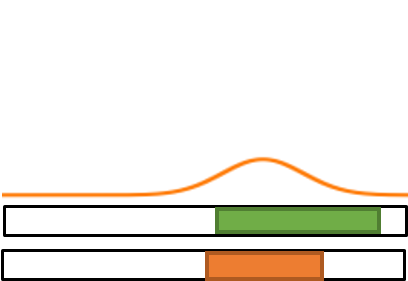}
		\end{minipage}
	}
	\centering
	\caption{Attending weight distribution along the sequence. The four bars represent the whole sequence area, and green blocks denote the ground truth regions of the corresponding characters. (a) Before extending, the blue attending region is narrow and prone to drift. (b) After extending, the brown attending region  is wider and has more overlapping with the ground truth region.
		This shows that the refined attention network outputs more reliable character features, which will benefit the following adversarial learning.
	}
	\label{fig:alpha}
\end{figure}

\emph{Inclusive attending process}.
Actually, it is difficult to directly apply the above character features to the following adversarial learning.
This is because 1) the \emph{attention drift} problem~\cite{cheng2017focus} causes the missing of key character features;
2) the attending region is too narrow to capture the features of the whole character area.
Specifically, in the attention-based character decoding process, the attending drift problem or the sharp attending region phenomenon is common, which may result in the missing of suboptimal character features. For example, in Fig.~\ref{fig:alpha}(a), only the blue bar is attended while the remaining character feature regions are lost.
These issues will impact the vulnerable adversarial learning.

To address this problem, we design an \textit{inclusive attending process} by extending the learnt attention area as follows:
\begin{equation}
\begin{split}
\alpha'_{t,j}      &=      \lambda\alpha_{t,j} +\frac{1-\lambda}{\eta(1+\eta)} \sum_{i=1}^{\eta}A(t,j-i)(\eta+1-i) \\
&\quad +\frac{1-\lambda}{\eta(1+\eta)} \sum_{i=1}^{\eta}A(t,j+i)(\eta+1-i)  \\
s.t. & ~~~~~A(t,j\pm i)=
\left\{	
\begin{aligned}
&\alpha_{t,j \pm i}  & 1\leq j \pm i \leq M \\
&\alpha_{t,j} & otherwise\\
\end{aligned}
\right.\\
\end{split}
\label{inclusive}
\end{equation}
where $\lambda \in [0, 1]$ is a tunable parameter that controls the decay degree of current attending weights. Specifically, $\lambda$ = 1 means there is no extending operation. And $\eta \in {\mathbb{Z}}^+$ controls the extending range.
Though the character features are the weight sum of the input features in the attention mechanism, in fact there are only a few input features that can impact the output.
The \textit{inclusive attending process} deals with this problem by re-weighting the attention weights.
$\alpha'_{t,j}$ is the re-weighted attention weight of the $t$-th character for $x_y$, $\left\{\alpha_{t,j \pm i}\right\}_{i=1}^\eta$ are the attention weights close to the $t$-th character, and the function $A(t,j \pm i)$ guarantees the legality of the output. $\alpha'_{t,j}$ is affected not only by the previous weight $\alpha_{t,j}$, but also by neighbors' weights. By this way, the attending region will be wider and more likely to cover the expected text in the image.
Meanwhile, The sum of attention weights is a constant~(equal to 1), which means that we do not need to normalize the weights after this process.
Fig.~\ref{fig:alpha}(b) illustrates the effect of the extending operation.
Thus, we replace the attending weight of Eq.~(\ref{eq:alpha}) by that of Eq.~(\ref{inclusive}).

Then, we capture the corresponding characters' features by training the attention model in a weakly-supervised way with only sequence-level annotations.
That is, each represented $CR_t^+$ is predicted as a target label $y_t=softmax(U^T CR_t^+)$ where $U$ is a trainable parameter.
Besides, the network need to process texts of variable lengths. Following~ \cite{sutskever2014sequence} , a special end-of-sentence (EOS) token is added
to the target set, so that the decoder completes the generation
of characters when EOS is emitted.
The loss function of the attention model is
\begin{equation}
	\mathcal{L}_{att}=-\sum_{t}logP(\hat{y}_t|\mathcal{I},\theta_{att})
	\label{eq:loss_att}
\end{equation}
where $\hat{y}_t$ is the ground truth of the $t\text{-}th$ character and $\theta_{att}$ is a vector of all trainable parameters of the attention network.



\subsection{Few-shot Adversarial Learning}\label{att_fada}
With the attending mechanism above, any text image can be encoded as a set $(CR_1, CR_2, ..., CR_L)$, where $L$ is the number of characters.
Now, we are to build the domain adaptation between the source domain $D_S$ and target domain $D_T$. 
As done in adversarial learning, we alternately learn a \emph{discriminator} and a character feature \emph{generator} as follows:

\emph{Discriminator learning}.
Given a represented source image $\mathcal{I}_s$ and a target image $\mathcal{I}_t$, FASDA first defines 4 categories of character pairs ($\mathcal{G}_i, i\in \{1,2,3,4\}$) between the source character set from $\mathcal{I}_s$ and the target character set from $\mathcal{I}_t$ for adversarial training.
Concretely, each character pair falls into one of the following four categories:
  1) Category 1 ($\mathcal{G}_1$): the two represented characters have the same class
   label and are sampled from the source character set;
  2) Category 2 ($\mathcal{G}_2$): the two represented characters have the same class label but are sampled separately from the source set and the target set;
  3) Category 3 ($\mathcal{G}_3$): the two represented characters have different class labels but both are sampled from the source set;
  4) Category 4 ($\mathcal{G}_4$): the two represented characters have different class labels and are sampled separately from the source set and the target set.

Then, we learn a \emph{multi-class discriminator} (denoted by MCD) in order to do semantic alignment of the source domain and the target domain. That is, MCD is optimized according to the standard cross-entropy loss as follows:
\begin{equation}
\mathcal{L}_{D}=-\sum_{i=1}^{4}\sum_{S \in \mathcal{G}_i}y_{\mathcal{G}_i}log(D(\phi(\mathcal{S})))
\label{eq:fada-d}
\end{equation}
where $y_{\mathcal{G}_i}$ is the group label of $\mathcal{G}_i$ and $D$ is the MCD function that consists of three fully connected layers. $\phi$ is a symbolic function that outputs the concatenation of a given feature pair.

\begin{figure}[t]
	\begin{center}
		\includegraphics[width=0.45\textwidth]{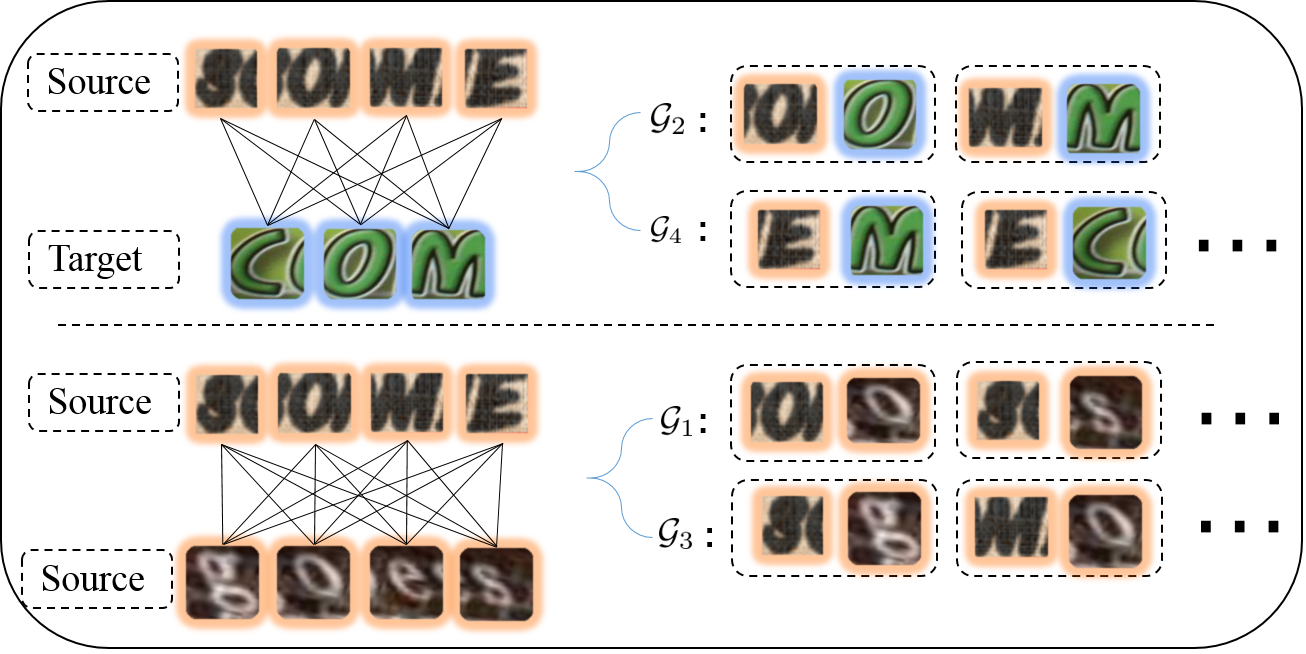}
	\end{center}
	\caption{Illustration of the sampling process.
		In the left, each row is a set of represented characters encoded from a text image of source or target domain. Pairs are generated by sampling from the source and target sets of represented characters, and are grouped into 4 categories ($\mathcal{G}_i$, $i$=1, 2, 3 and 4) in the right.
	}
	\label{fig:sample}
\end{figure}

Different from the existing sampling methods that sample examples from the global (image-level) categories, here we sample pairs from character-level categories, i.e., two encoded unique-character sets.
The character-level \emph{sampling strategy} for generating those pairs is illustrated in Fig.~\ref{fig:sample} and detailed in Algorithm \ref{alg:generate}. Here, Line 1 encodes images to character representations, Lines 2 and 3 generate represented character features pairs of categories $\mathcal{G}_2$ and $\mathcal{G}_4$, $\mathcal{G}_1$ and $\mathcal{G}_3$. Finally, Line 4 collects all pairs.
\begin{algorithm}[h]
	\centering
	\scriptsize
	\caption{The sampling strategy.}
	\label{alg:generate}
	\renewcommand{\algorithmicrequire}{\textbf{Input:}}
	\renewcommand{\algorithmicensure}{\textbf{Output:}}
	\begin{algorithmic}[1]
		\REQUIRE ~~\\
				\textbf{Input:}
		A source image $\mathcal{I}^s$ and its labels $\hat{\mathcal{Y}}^s=(\hat{y}_1^s, \dots, \hat{y}_L^s)$, \\
		A target image $\mathcal{I}^t$ and its labels $\hat{\mathcal{Y}}^t=(\hat{y}_1^t, \dots, \hat{y}_{L^\prime}^t)$.
		
		\ENSURE ~~\\
		$\mathcal{G}_i, i\in \{1,2,3,4\}$; \\
		\STATE Encoding $\mathcal{I}^s$ and $\mathcal{I}^t$ as
		$CR^s=(CR_1^s, \dots, CR_L^s)$ and
		$CR^t=(CR_1^t, \dots, CR_L^t)$, respectively.\\
		
		\STATE Sampling all pairs of represented characters $CR^s \times CR^t$ to obtain $\mathcal{S}_2$ or $\mathcal{S}_4$:
		\begin{equation}
		\mathcal{S}_k = \left\lbrace CR_i^s, CR_j^t \right\rbrace,\nonumber
		\end{equation}
		where $k$=2 if $\hat{y}_i^s$ = $\hat{y}_j^t$, otherwise $k$=4.
		
		\STATE Sampling all pairs from $CR^s \times CR^s$ to obtain $\mathcal{S}_1$ or $\mathcal{S}_3$:
		\begin{equation}
		\mathcal{S}_k = \left\lbrace CR_i^s, CR_j^s \right\rbrace,\nonumber
		\end{equation}
		\label{code:fram:add}
		where $k$=1 if $\hat{y}_i^s$ = $\hat{y}_j^s$, otherwise $k$=3.
		
		\STATE Collecting all pairs $\mathcal{S}_i$ into $\mathcal{G}_i,i\in \{1,2,3,4\}$.
	\end{algorithmic}
\end{algorithm}

\emph{Generator learning}.
In this stage, we update the attention network to confuse MCD so that it can no longer distinguish $\mathcal{G}_1$ and $\mathcal{G}_2$ or $\mathcal{G}_3$ and $\mathcal{G}_4$.
Thus, the generator is learnt by
\begin{equation}
\mathcal{L}_{G}=-\left[\sum_{\mathcal{S} \in \mathcal{G}_2} y_{\mathcal{G}_1}log(D(\phi(\mathcal{S})))+
\sum_{\mathcal{S} \in \mathcal{G}_4} y_{\mathcal{G}_3}log(D(\phi(\mathcal{S})))\right].
\label{eq:fada-g}
\end{equation}
This loss function forces the attention network to embed target samples into a space such that MCD cannot distinguish the source and target, which greatly promotes the process of domain confusion and the semantic alignment of classes.

However, the adversarial learning above tries only to make the source domain and the target domain indistinguishable, but does not consider the context information among characters in the represented character sequences.
This means that we should also optimize $\mathcal{L}_{att}$ to remain the context information.
To this end, the generator considers both $\mathcal{L}_{G}$ and $\mathcal{L}_{att}$. Formally, we have
\begin{equation}
\begin{split}
\mathcal{L}_{Att-G} = \gamma\mathcal{L}_{G} + \mathcal{L}_{att}
\end{split}
\label{eq:attfada}
\end{equation}
where $\gamma$ is for controlling the impact of confusion.

\subsection{Training Process}\label{sec:training}
As in most adversarial learning, we alternately train $\mathcal{L}_{Att-G}$ and $\mathcal{L}_{D}$ to update the attention network and MCD.
We first pre-train the attention network by optimizing $\mathcal{L}_{att}$ using the labeled source data, then create groups by using the proposed sampling algorithm (Alg.~\ref{alg:generate}) for pre-training MCD.
In the \emph{Discriminator learning} stage, we freeze the parameters of the attention network for optimizing MCD, while in the \emph{Generator learning} stage, we freeze the parameters of MCD for optimizing the attention network.

\section{Performance Evaluation}

\subsection{Datasets and Implementation Details}
{\bf{Datasets}}: We use 8 million synthetic text image data (refer to as Synthetic)~\cite{jaderberg2014synthetic} as the source domain, and select some popular benchmarks as target domains, including
{\bf{Street View Text}} (\emph{abbr.} SVT)~\cite{wang2011end},
{\bf{ICDAR 2003}} (\emph{abbr.} IC03)~\cite{lucas2003icdar},
{\bf{ICDAR 2013}} (\emph{abbr.} IC13)~\cite{karatzas2013icdar}, and
{\bf{ICDAR 2015}} (\emph{abbr.} IC15)~\cite{karatzas2015icdar}.
%
%
For sufficiently evaluating the domain adaptation performance of our method, we also collect several specific datasets, including one dataset of licence plates (denoted by `A'), one dataset of container numbers (denoted by `B'), one dataset of batch numbers (denoted by `C'), and one dataset collected from assembly lines (denoted by `D').
Table \ref{tab:datasets} gives the details of these datasets.

\begin{table}[!th]
	\begin{center}	
		\caption{Statistics of synthetic, public and specific datasets. `K' and `M' separately mean thousand and million.}
		\label{tab:datasets}	
		\scalebox{0.75}{
			\begin{tabular}{|l|c|cccc|cccc|}
				\hline
				\multirow{2}{*}{Dataset}&\multirow{2}{*}{Synthetic}&\multicolumn{4}{c|}{Public}&
				\multicolumn{4}{c|}{Specific}\cr
				\cline{3-10}
				&~& SVT&IC03&IC13&IC15&A&B&C&D\cr
				\hline
				Train &8M&257&936&680&4066&5K&5K&5K&5K \cr
				Test&-&647&867&1015&1811&5K&5K&5K&5K\cr
				\hline
			\end{tabular}
		}
	\end{center}
\end{table}


{\bf{Network Structure}}:
The extractor of our method is inherited from the network architecture (7 convolution layers and 1 LSTM layer) used in \cite{shi2016robust}.
The attention network is implemented with
an LSTM (256 memory blocks) and 37 output units (26 letters,
10 digits, and 1 end-of-symbol).
The MCD contains 3 additional fully connected layers with 1024, 1024 and 4 dimensions, respectively. 

{\bf Training details}:
The attention network is pre-trained by ADADELTA~\cite{adadelta} with source datasets.
The adversarial learning process is optimized by Adam~\cite{adam} with the learning rate being 0.001.
The batch size is set to 64, and images are scaled to $256 \times 32$ in both training and testing.
$\gamma$ in Eq.~(\ref{eq:attfada}) is 0.00005,
$\eta$ in Eq.~(\ref{inclusive} ) is 1.0, and $\lambda$ in Eq.~(\ref{inclusive}) is 0.75.
Besides, $\lambda$ and $\eta$ are selected by conducting the parameter traversal strategy. $\gamma$ is an empirically value that is set stable for model training. As for the learning rate and batch size, we just follow the previous STR works.

\subsection{Domain Adaptation on Public Benchmarks} \label{sec:ablation}
For evaluating our method, we randomly select 20-thousand images from the synthetic dataset (described in Table~\ref{tab:datasets}) as the source domain, and randomly choose 150
images from each public benchmark (\emph{e.g.} IC03) as the target domain.

To be fair, we provide three settings as \emph{baselines}:
\emph{Setting 1}: Testing benchmarks with the model trained on the source dataset, denoted as ``Source Only'';
\emph{Setting 2}: Finetuning with only 150 selected target samples (shown in Tab. \ref{tab:datasets}) and then  conducting testing, denoted as ``FT \emph{w/} T'';
\emph{Setting 3}: Finetuning with all source samples and the available target samples (the ratio of source/target is 20 in each batch), denoted as ``FT \emph{w/} S+T''. Note that the ratio of source samples over target samples has been well-tuned in our experiments.
\begin{table}[h]
	\centering	
	\caption{
		Domain adaptation results on four benchmarks. ``IA'' means the model with the \textbf{i}nclusive \textbf{a}ttention mechanism.
	}
	\label{tab:special_dataset}
	\begin{tabular}{lcccc}
		\toprule
		Method  & SVT & IC03 & IC13 & IC15 \\
		\midrule
		Source Only     & 19.6  & 44.1  & 46.8  & 14.5  \\
		FT \emph{w/} T  & 23.9 &46.9&49.7&15.5 \\
		FT \emph{w/} S+T  & 25.1  & 52.3  & 51.1  & 16.4   \\
		FASDA-$CR$ & 27.5 & 55.8 & 54.9 & 18.6 \\
		FASDA-$CR^+$  &  28.8 & 56.8  & 56.6  & 19.1   \\
		FASDA-IA-$CR^+$  &  {\bf 29.4} & {\bf 58.1}  & {\bf 57.5}  & {\bf 19.2}   \\
		\bottomrule
	\end{tabular}
\end{table}
\begin{table}[!th]
	\begin{center}			
		\caption{
			Domain adaptation results from synthetic domain to specific domains. ``10'', ``100'', ``1000'' and ``3000'' indicate the size of target samples for domain adaptation.
		}
		\label{tab:results-synth}
		\scalebox{0.98}{
			\begin{tabular}{llcccc}
				\hline
				&                 &10     &100   &   1000 & 3000 \cr \hline
				\multirow{2}{*}{\small{Synthetic$\rightarrow$A}}  & FT \emph{w/} S+T  &8.9	 &24.3  &	71.4 &	89.8 \cr
				& FASDA             &\textbf{10.2}   &\textbf{32.5}  &	\textbf{78.7} & \bf{	92.9} \cr \hline
				
				\multirow{2}{*}{\small{Synthetic$\rightarrow$B}}  & FT \emph{w/} S+T  &1.3&	6.8	&43.9&	59.6 \cr
				& FASDA             &\textbf{2.9}&	\textbf{13.6}&	\textbf{48.2}&	\bf{62.7} \cr \hline
				\multirow{2}{*}{\small{Synthetic$\rightarrow$C}}  & FT \emph{w/} S+T  &1.8	&7.5&	55.4&	70.0 \cr
				& FASDA             &1.8	&\textbf{11.9}&	\textbf{59.1}& \bf{74.9} \cr \hline
				
				\multirow{2}{*}{\small{Synthetic$\rightarrow$D}}  & FT \emph{w/} S+T  &0.0&	2.4&	48.6&	71.0 \cr
				& FASDA             &0.0&	\textbf{6.3}&	\textbf{64.}1&	\bf{76.2} \cr \hline
			\end{tabular}
		}
	\end{center}
\end{table}

Table~\ref{tab:special_dataset} summaries the results.
We can see that 1) our method significantly outperforms all baselines on all benchmarks, which validates the effectiveness of the FASDA method.
2) Sequence domain adaptation with contextual character representation ($CR^+$) can achieve better performance.
3) IA devotes to enlarging/smoothing the attending range, and the outstanding performance of FASDA-IA-$CR^+$ shows the effect of IA.
%
%
So without specific declaration, we use FASDA as FASDA-IA-$CR^+$ in the following experiments.

\subsection{Domain Adaptation on Specific Datasets}
\begin{table*}[t]
\begin{center}	
	\caption{
		Domain adaptation results from specific domains to specific domains. Here, ``100'' and ``1000'' are the number of target samples.
	}
	\label{tab:results-cross}
	\scalebox{0.95}{
	\begin{tabular}{llcccccccccccc}
		\toprule
		&           &A$\rightarrow$B &A$\rightarrow$C &A$\rightarrow$D
		&B$\rightarrow$A &B$\rightarrow$C &B$\rightarrow$D
		&C$\rightarrow$A &C$\rightarrow$B &C$\rightarrow$D
		&D$\rightarrow$A &D$\rightarrow$B &D$\rightarrow$C \cr
		\midrule
		\multirow{2}{*}{100}  & FT \emph{w/} S+T  	 &0.0 &0.0&0.0&16.6&9.9&21.9&5.2&0.5&10.2&14.4&0.3&6.1  \\
		& FASDA         &0.0&0.0&0.0&\textbf{26.5}&\textbf{35.8}&\textbf{34.7}&\textbf{13.3}&\textbf{8.4}&\textbf{30.8}&\textbf{18.9}&\textbf{7.8}&\textbf{19.2}  \\
		\midrule
		\multirow{2}{*}{1000}  & FT \emph{w/} S+T     &0.2&30.0&13.8&68.8&67.9&59.5&52.6&41.5&56.4&54.5&38.3&62.5 \\
		& FASDA         &\textbf{45.1}&\textbf{78.9}&\textbf{69.7}&\textbf{85.0}&\textbf{78.0}&\textbf{70.9}&\textbf{75.3}&\textbf{60.4}&\textbf{72.8}&\textbf{78.1}&\textbf{63.7}&\textbf{80.9} \\
		\bottomrule
	\end{tabular}
}
\end{center}
\end{table*}

To further evaluate the domain adaptation performance of our method, we evaluate our method on four collected specific datasets.
Concretely, we conduct two kinds of domain adaptation: \emph{synthetic $\dashrightarrow$ specific} as \emph{Case 1} and \emph{specific $\dashrightarrow$ specific} as \emph{Case 2}.

In \emph{Case 1}, the source contains 20 thousands synthetic images randomly selected from \cite{jaderberg2014synthetic}, and each specific dataset is constructed by randomly selecting a certain number of images (\emph{e.g.} 10, 100, 1000 and 3000) from its corresponding training set (See Table \ref{tab:datasets}). Then, we conduct testing on the test datasets, each of which contains 5000 images (See Table \ref{tab:datasets}). In \emph{Case 2}, the sources are the corresponding specific training sets (5000 images, see Table \ref{tab:datasets}), while the construction of target datasets and testing datasets is similar to that of \emph{Case 1}.

The results of \emph{Case 1} and \emph{Case 2} are separately given in Table~\ref{tab:results-synth} and Table~\ref{tab:results-cross}. We can see that FASDA \emph{always achieves better} results than FT \emph{w/} S+T in all settings,
which shows that our method can work well on real scenarios.
Of course, we also see that 1) the gap between FASDA and FT \emph{w/} S+T gradually decreases with the increase of target domain dataset size. Even so, FASDA still outperforms FT \emph{w/} S+T.
2) Some 0\% accuracy values appear in Table~\ref{tab:results-synth} and \ref{tab:results-cross}. There are two possible reasons: one is the gap between the source and target domains, the other is that there are too few training samples to obtain a good model. In summary,
	the observation above indicates the necessity of sequence-level domain adaptation for STR, especially on real datasets.

\subsection{Comparison with the state-of-the-art}
\begin{table*}[t]
	\begin{center}	
		\caption{Results of domain adaptation from a large source to public benchmarks. ``50'' is the lexicon size. ``Full'' indicates the combined lexicon of all images in the benchmarks. ``None'' means lexicon-free.
}
		\label{tab:results-regular2}
			\begin{tabular}{|l|cc|ccc|c|c|}
				\hline
				\multirow{2}{*}{{Method}} & \multicolumn{2}{c|}{{SVT}} & \multicolumn{3}{c|}{{IC03}} & {IC13}& {IC15}  \cr\cline{2-8}
				& {50} & {None}  & {50} & {Full} & {None} & {None} & {None}  \cr
				\hline
				Yao \etal(2014)\cite{yao2014strokelets}&75.9&-&88.5&80.3&-&-&-\cr
				Jaderberg \etal(2016)\cite{jaderberg2016reading} & 95.4& 80.7& 98.7& \bf{98.6}& 93.1& 90.8& - \cr
				Shi \etal(2017)\cite{shi2016end} & 96.4& 80.8& 98.7& 97.6& 89.4& 86.7& - \cr
				Lee\&Osindero~(2016)\cite{lee2016recursive} & 96.3& 80.7& 97.9& 97.0& 88.7 &90.0& - \cr
				Cheng \etal(2018)\cite{cheng2018aon}& 96	&82.8&	98.5&	97.1&	91.5&	-&	68.2 \cr
				Bai \etal(2018)\cite{bai2018cvpr}&	96.6&	87.5&	98.7&	97.9&	94.6&	\bf{94.4}&	73.9 \cr
				Liu \etal(2018)\cite{liu2018synthetically}&96.8&	87.1&	98.1&	97.5&	94.7&	94.0 & - \cr
				Shi \etal(2018)\cite{shi2018aster}&\bf{99.2}&	\bf{93.6}&	\bf{98.8}&	98.0&	94.5&	91.8&	76.1\cr
				Li \etal(2019)\cite{li2019strongbaseline}&98.5&	91.2&-&-&-&				94.0&	\bf{78.8}\cr
				Luo \etal(2019)\cite{luo2019moran}&96.6&	88.3&	98.7&	97.8&	\bf{95.0}&	92.4&	68.8\cr
				Zhang \etal(2019)\cite{zhang2019sequence}&-&	84.5&	-&	-&	92.1&	91.8&	- \cr
				\hline
				Shi \etal(baseline)(2016)\cite{shi2016robust} &96.1& 81.5 &97.8& 96.4& 88.7& 87.5& - \cr
				Cheng \etal(baseline)(2017)\cite{cheng2017focus} &95.7& 82.2 &98.5& 96.7& 91.5& 89.4& 63.3 \cr
				Shi \etal(baseline)(2018)\cite{shi2018aster} &-&	\textbf{91.6}&	-&	-&	93.6&	90.5&	- \cr
				Luo \etal(baseline)(2019)\cite{luo2019moran} &-&	84.1&	-&	-&	92.5&	90.0&	68.8 \cr
				\hline
				Source Only &\textbf{96.8}&	85.2&	99.0&	97.5&	92.3&	91.6&	68.2 \cr
				FT \emph{w/} S+T &96.4&	86.5&	98.7&	\textbf{97.6}&	93.0&	92.4&	71.8 \cr
				FASDA& 96.5&	88.3&	\textbf{99.1}&	97.5&	\textbf{94.8}&	\textbf{94.4}&	\textbf{73.3} \cr
				\hline
			\end{tabular}
	\end{center}
\end{table*}

We also evaluate FASDA with a strong ResNet-based feature extractor released by \cite{cheng2017focus}, and conduct domain adaptation from a large source domain (12 million synthetic images used in \cite{cheng2017focus}) to several public benchmarks with all training targets (their sizes are given in Table~\ref{tab:datasets}). 

As there are not ancillary strategies such as character-level annotations in our method.
To be fair, we compare our method with the baselines of some state of the art works. All results are given in Table~\ref{tab:results-regular2}.
We can see that our method obviously outperforms the four baselines in almost all cases, except for Shi \etal~(baseline)(2018) on SVT where a bidirectional decoder was used.
Furthermore,
we can also see that FASDA performs clearly better than FT \emph{w/} S+T almost on all benchmarks.
Here, note that in real applications, lexicon-free performance is more desirable than that with lexicon. And for any method, its lexicon-free performance does not always correspond to that with lexicon, which was also observed in many existing works \cite{cheng2017focus,luo2019moran}.
Therefore, FASDA can serve as a \emph{general performance boosting strategy} in text recognition tasks.

\subsection{Effect of IA}
Here we treat the features generated by the attention model as the corresponding character features, but this may be incorrect due to the \emph{attention drift} problem~\cite{cheng2017focus}.
Therefore, we design the IA process to handle this problem.
Here, we evaluate the effectiveness of IA by calculating the Character Generation Accuracy (CharAcc in short) as follows:
\begin{equation}
CharAcc = \frac{\mathbf{N}(y,gt)}{|gt|}
\label{eq:cha_acc}
\end{equation}
where $y$ is the predicted character string and $gt$ is the character sequence ground truth, and the function $\mathbf{N}()$ returns the number of correctly recognized characters (counting by aligning $y$ and $gt$).
In some sense, when a character is recognized, it is mainly because the attention model focuses on the right position. So
if IA works, the value of CharAcc should be large. 
Table \ref{tab:character_dataset} shows the results.

From Table \ref{tab:character_dataset} we can see that on
SVT, IC03 and IC13, the introduction of IA does boost the reliability of character features, which helps the adversarial few-shot learning. However, on
IC15 we can see that FASDA performs worse than FT.
This is because IC15 is so irregular that the attention model cannot properly locate the characters.
By comparing the results of with and without IA (the last two rows in Table~\ref{tab:character_dataset}), we can see that the IA process is definitely able to improve the character recognition performance.

\begin{figure}[t!]
	\begin{center}
		\includegraphics[width=0.40\textwidth]{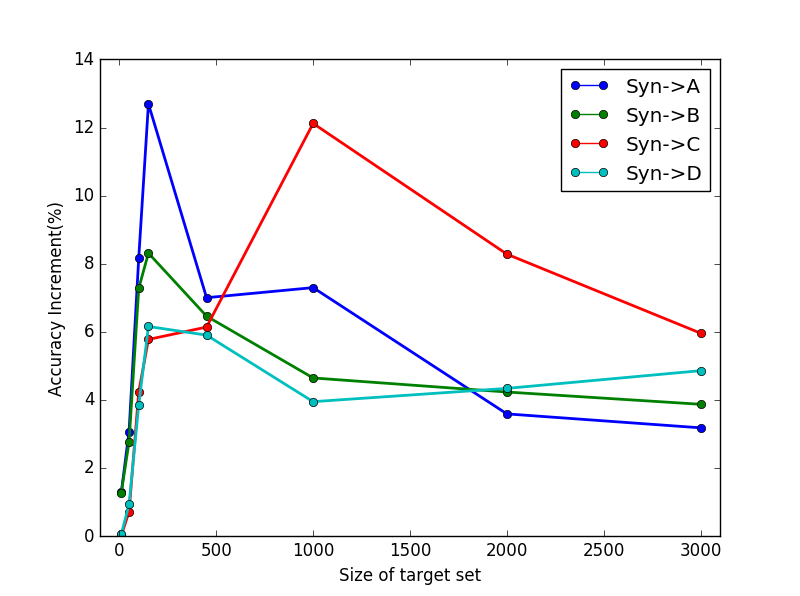}
	\end{center}
	\caption{Accuracy increment \emph{vs}. target set size.}
	\label{fig:targetsize}
\end{figure}

\subsection{Effect of Target Size}
Here we evaluate the effect of target training set size.
Due to the limited training data of public benchmarks, we conduct experiments on four collected specific datasets.
We use \emph{Accuracy Increment} of our method over FT \emph{w/} S+T to measure the effect of target size, which is evaluated by (\emph{accuracy of FASDA} - \emph{accuracy of FT \emph{w/} S+T}). The same settings as in Table~\ref{tab:special_dataset} is used.
Fig.~\ref{fig:targetsize} shows the results, from which we can see the following general trend: as target set size increases, the \emph{accuracy increment} first rapidly increases, and after reaching the maximum it then decreases gradually. 
	For small target set (size $\le$ 10), there are too few samples in the target set, and for some classes there are not even any sample. So the \emph{accuracy increment} is low due to the lack of target samples.
	For middle target set (size $>$10 but $\le$ 1000), the \emph{accuracy increment} gradually increases with the target set size. Though the size of the target set is still too few to finetune a good model that can perform very well in the target domain, the performance of the proposed method turns better as more target samples are used in model training.
	And for large target set (size $>$1000), the \emph{accuracy increment} gradually decreases or maintains stable as the target samples are already sufficient enough. Whatever, FASDA outperforms FT w/ S+T.


\begin{table}[h]
	\centering	
\caption{
	Character generation accuracy on four benchmarks.
}
	\label{tab:character_dataset}
	\begin{tabular}{lcccc}
		\toprule
		Method  & SVT & IC03 & IC13 & IC15 \\
		\midrule
		Source Only     & 51.9 & 68.8 & 66.4 & 36.8  \\
		FT \emph{w/} T  & 56.6 & 72.4 & 68.5  & 42.8 \\
		FT \emph{w/} S+T  & 58.0 & 74.7 & 70.4 & \textbf{43.1}   \\
		\hline
		FASDA-$CR$  & 58.1 & 75.5 & 73.0 & 39.6 \\
		FASDA-$CR^+$   & 58.9 & 76.3 & 73.6 & 40.1   \\
		FASDA-IA-$CR^+$  & \textbf{59.9 }& \textbf{76.9} & \textbf{74.3} & 41.3   \\
		\bottomrule
	\end{tabular}
\end{table}

\section{Conclusion}
In this paper, we introduce a few-shot adversarial sequence domain adaptation (FASDA) approach to implement sequence-level domain adaptation for STR.
The proposed method first applies a well-designed attending mechanism to represent each character's feature, then conducts domain adaptation by adversarial learning. The new method can maximize the character-level confusion between the source domain and the target domain, and thus achieves good performance in real scenarios even when there are only a few labeled samples.
In the future, we plan to further optimize the sequence-level domain adaptation method.







\bibliographystyle{IEEEtran}
\bibliography{ijcai19}
%

\end{document}